\pdfoutput=1

\documentclass[11pt]{article}

\usepackage[]{naacl2021}

\usepackage{times}
\usepackage{latexsym}

\usepackage[T1]{fontenc}

\usepackage[utf8]{inputenc}

\usepackage{microtype}

\usepackage{amsmath,amsfonts,amssymb,bm}
\usepackage{pifont} 
\usepackage{graphicx,subfigure,epsfig,fancybox} 
\usepackage{float}
\usepackage{color} 
\usepackage{multirow}
\usepackage{natbib}

\usepackage{tikz}
\usetikzlibrary{shapes.geometric, positioning, calc}
\usetikzlibrary{arrows,shapes,calc}
\usetikzlibrary{trees,positioning,mindmap,shadows,fit}
\usetikzlibrary{decorations.pathreplacing}
\usetikzlibrary{tikzmark} 
\usetikzlibrary{intersections} 

\newcommand{\revised}[1]{\textcolor{blue}{}}
\renewcommand{\vec}[1]{\boldsymbol{#1}}

\newcommand{\expect}[2]{\mathbb{E}_{#2}[{#1}]}

\usepackage{adjustbox}
\usepackage{array}
\usepackage{booktabs}

\newcolumntype{R}[2]{%
    >{\adjustbox{angle=#1,lap=\width-(#2)}\bgroup}%
    l%
    <{\egroup}%
}

\usepackage{setspace}

\usepackage{trackchanges}
\addeditor{YJ}
\addeditor{Hanjie}

\usepackage{cleveref}
\usepackage{bbm}
\usepackage{subfigure}
\usepackage{xcolor}
\usepackage{soul}
\usepackage{xspace}

\newcommand{\hlc}[2][yellow]{{%
		\colorlet{foo}{#1}%
		\sethlcolor{foo}\hl{#2}}%
}

\newcommand{\ourmethod}{\textsc{GMASK}\xspace}

\crefformat{section}{\S#2#1#3}
\crefformat{subsection}{\S#2#1#3}
\crefformat{subsubsection}{\S#2#1#3}

\newcommand*{\affaddr}[1]{#1} 
\newcommand*{\affmark}[1][*]{\textsuperscript{#1}}
\newcommand*{\email}[1]{\small\texttt{#1}}

\title{Explaining Neural Network Predictions on Sentence Pairs via Learning Word-Group Masks}

\author{%
	Hanjie Chen\affmark[1]\quad Song Feng\affmark[2]\quad Jatin Ganhotra\affmark[2]\quad Hui Wan\affmark[2] \\
	\textbf{Chulaka Gunasekara\affmark[2]\quad Sachindra Joshi\affmark[2]\quad Yangfeng Ji\affmark[1]}\\
	\affaddr{\affmark[1]Department of Computer Science, University of Virginia, Charlottesville, VA, USA}\\
	\affaddr{\affmark[2]IBM Research AI}\\
	\email{\{hc9mx, yangfeng\}@virginia.edu}\\
	\email{\{sfeng, jatinganhotra, hwan\}@us.ibm.com}\\
	\email{\{Chulaka.Gunasekara@, jsachind@in.\}ibm.com}\\
}

\begin{document}
\maketitle

\begin{abstract}
	Explaining neural network models is important for increasing their trustworthiness in real-world applications. 
Most existing methods generate post-hoc explanations for neural network models by identifying individual feature attributions or detecting interactions between adjacent features.
However, for models with text pairs as inputs (e.g., paraphrase identification), existing methods are not sufficient to capture feature interactions between two texts and their simple extension of computing all word-pair interactions between two texts is computationally inefficient. 
In this work, we propose the Group Mask (\ourmethod) method to implicitly detect word correlations by grouping correlated words from the input text pair together and measure their contribution to the corresponding NLP tasks as a whole.
The proposed method is evaluated with two different model architectures (decomposable attention model and BERT) across four datasets, including natural language inference and paraphrase identification tasks. 
Experiments show the effectiveness of \ourmethod in providing faithful explanations to these models \footnote{Code for this paper is available at \url{https://github.com/UVa-NLP/GMASK}}.
\end{abstract}

\section{Introduction}
\label{sec:intro}

Explaining deep neural networks is critical for revealing their prediction behaviors and enhancing the trustworthiness of applying them in real-world applications. 
Many methods have been proposed to explain neural network models from the post-hoc manner that generates faithful explanations based on model predictions \citep{ribeiro2016should, lundberg2017unified, sundararajan2017axiomatic, guidotti2018survey}. 
Most existing work focuses on identifying word attributions \citep{rocktaschel2015reasoning, li2016understanding, thorne2019generating} for NLP tasks.
Knowing which individual features are important might not be enough for explaining model behaviors.
Then, other recent work exploits feature interactions as explanations \citep{singh2018hierarchical, chen2020generating, tsang2020feature}.
However, they could suffer computation inefficiency while computing interactions between all word pairs, and they also fall short for identifying multiple important words correlated from different input sources for predictions.
Such intuitions are particularly important for explaining sentence pair modeling tasks such as natural language inference (NLI) \citep{bowman2015large} and paraphrase identification (PI) \citep{yin2015convolutional}.

\begin{figure}[t]
  \centering
  \includegraphics[width=0.45\textwidth]{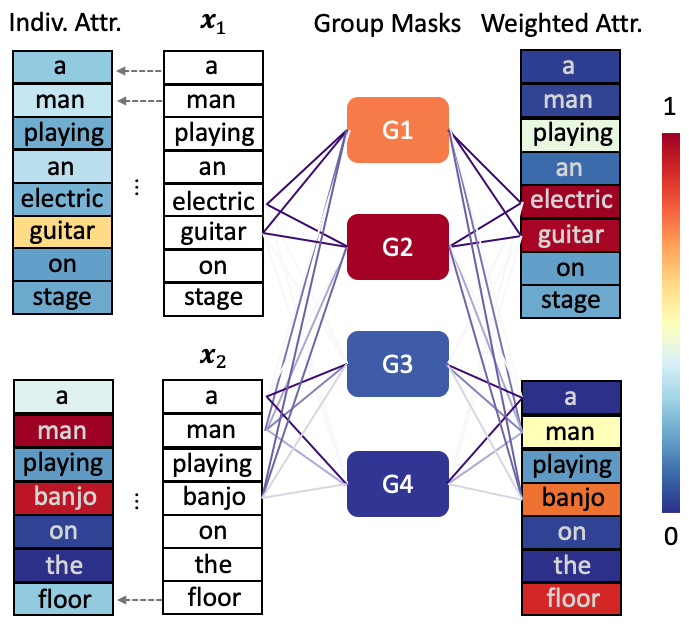}
  \caption{\label{fig:illustrations} An illustration of obtaining individual word attributions (Indiv. Attr.) and weighted word attributions (Weighted Attr.), where the color of each block represents word importance or group importance, and the color saturation of purple lines indicates the probability of a word belonging to a specific group.}
\end{figure}

\autoref{fig:illustrations} shows an example of NLI, where the model makes correct prediction as \textsc{contradiction}. 
The first column visualizes individual word attributions to the prediction, where the top four important words are \texttt{man}, \texttt{banjo}, \texttt{guitar}, \texttt{a}. However, the correlations between them are unclear and intuitively \texttt{man} and \texttt{a} are irrelevant to the model prediction, which makes the explanation untrustworthy. A good explanation should be able to capture correlated words between the sentence pair, and identify their importance to the model prediction.

In this work, we propose Group Masks (\ourmethod), a model-agnostic approach that considers the importance of correlated words from two input sentences. In particular, it distributes correlated words into a group and learns the group importance. In \autoref{fig:illustrations}, the input words are distributed in four groups with importance $G2>G1>G3>G4$. The color saturation of purple lines represents the probability of a word belonging to a group. Different from individual word attributions, \ourmethod assigns \texttt{electric}, \texttt{guitar}, and \texttt{banjo} into important groups ($G2/G1$), while \texttt{man} and \texttt{a} into unimportant groups ($G3/G4$). The weighted word attributions computed as the weighted sum of group importance identify the important words \texttt{electric}, \texttt{guitar} from $\vec{x}_1$ and \texttt{banjo} from $\vec{x}_2$, which explains the model prediction.

The contribution of this work is three-fold: (1) we introduce \ourmethod method to explain sentence pair modeling tasks by learning weighted word attributions based on word correlations; (2) we propose a sampling-based method to solve the optimization objective of \ourmethod; and (3) we evaluate the proposed method with two types neural network models (decomposable attention model \citep{parikh2016decomposable} and BERT~\cite{devlin2018bert}), for two types of sentence pair modeling tasks on four datasets. Experiments show the superiority of \ourmethod in generating faithful explanations compared to other competitive methods.

\section{Related Work}
\label{sec:relate}
Many approaches have been proposed to explain deep neural networks from the post-hoc manner, such as gradient-based explanation methods \citep{hechtlinger2016interpretation,sundararajan2017axiomatic}, attention-based methods \citep{ghaeini2018interpreting, serrano2019attention}, and decomposition-based methods \citep{murdoch2018beyond, du2019attribution}. However these white-box explanation methods are either rendering doubt regarding faithfulness \citep{jain2019attention, wiegreffe2019attention} or being limited to specific neural networks. In this work, we mainly focus on model-agnostic explanation methods, which are applicable to any black-box models.

\paragraph{Feature attributions}
Many approaches explain models by assigning feature attributions to model predictions. For example, perturbation-based methods quantify feature attributions by erasing features \citep{li2016understanding} or using local linear approximation as LIME \citep{ribeiro2016should}. KernelSHAP \citep{lundberg2017unified} utilized Shapley values \citep{shapley1953value} to compute feature attributions. Another line of work proposed learning feature attributions, such as L2X \citep{chen2018learning} which maximizes mutual information to recognize important features, and IBA \citep{schulz2020restricting} which identifies feature attributions by optimizing the information bottleneck \citep{tishby2000information}. However, these approaches produce individual word attributions without considering feature correlations. Our proposed method implicitly detects correlated words and generates weighted word attributions.

\paragraph{Feature interactions}
Some work proposed to generate explanations beyond word-level features by detecting feature interactions. \citet{murdoch2018beyond} proposed contextual decomposition (CD) to compute word interactions and \citet{singh2018hierarchical} and \citet{jin2019towards} further proposed hierarchical versions based on that. Other work adopted Shapley interaction index to compute feature interactions \citep{lundberg2018consistent} and build hierarchical explanations \citep{chen2020generating}. However, computing feature interactions between all word pairs is computationally inefficient \citep{tsang2018can}. Methods which only consider the interactions between adjacent features are not applicable to sentence pair modeling tasks as critical interactions usually form between words from different sentences. \ourmethod distributes correlated words from the input text pair into a group, and learns the group importance, without explicitly detecting feature interactions between all word pairs.

\paragraph{Word masks}
Some related work utilized word masks to select important features for building interpretable neural networks \citep{lei2016rationalizing, bastings-etal-2019-interpretable} or improving the interpretability of existing models \citep{chen2020learning}. \citet{de2020decisions} proposed to track the information flow of input features through the layers of BERT models. Different from the prior work, \ourmethod applies masks on a group of correlated words.

\section{Method}
\label{sec:method}
\begin{figure*}[htbp!]
  \centering
  \includegraphics[width=0.8\textwidth]{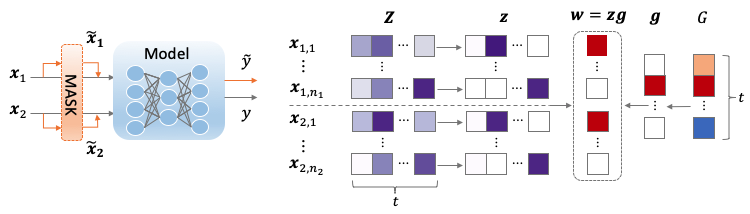}
  \caption{\label{fig:framework} The left part shows that masks are applied on the word embedding layer, selecting important words for the neural network model. The outputs $y$ and $\tilde{y}$ are corresponding to the original input $\vec{x}=[\vec{x}_1, \vec{x}_2]^{T}$ and masked input $\vec{\tilde{x}} = [\vec{\tilde{x}}_1, \vec{\tilde{x}}_2]^{T}$ respectively. The right part shows the sampling process of \ourmethod. For $\vec{Z}$, the color saturation of purple blocks represents the probability of a word belonging to a specific group (i.e. $\phi_{i, j}(\iota)$). $\vec{z}$ is a sample of $\vec{Z}$ with binary values. For $G$, the color of each block represents group importance. $\vec{g}$ is a one-hot vector sampled from $G$, indicating which group being selected. $\vec{w}$ is a sample of word masks obtained by multiplying $\vec{z}$ and $\vec{g}$.}
\end{figure*}

This section introduces the proposed \ourmethod method. \ourmethod implicitly learns word correlations, and distributes correlated words from different input sentences into a group. \ourmethod learns the importance of each group by randomly masking out groups of words. Finally, the weighted word attributions are computed based on word group distributions and group importance.

\subsection{Explaining Models with Word Masks}
\label{subsec:wmask}
As the left part of \autoref{fig:framework} shows, the word masks are applied on input word embeddings, learning to select important words to explain the model prediction. For each input data, we generate a post-hoc explanation by learning a set of mask values which represent the word attributions.

For sentence pair modeling tasks, the input contains two sentences $\vec{x}_{1}=[\vec{x}_{1, 1}^{T}, \dots, \vec{x}_{1, n_1}^{T}]^{T}$ and $\vec{x}_{2}=[\vec{x}_{2, 1}^{T}, \dots, \vec{x}_{2, n_2}^{T}]^{T}$, where  $\vec{x}_{i, j}\in \mathbb{R}^{d}$ ($i \in \{1, 2\},~j \in \{1, \dots, n_{i}\}$) represents the word embedding and $n_1$ and $n_2$ are the number of words in the two texts respectively. We denote the neural network model as $f(\cdot)$ which takes $\vec{x}_1$ and $\vec{x}_2$ as input and outputs a prediction label $y=f(\vec{x})$, where $\vec{x} = [\vec{x}_1, \vec{x}_2]^{T}$. To explain the model prediction, we learn a set of word masks $\vec{W}=[W_{1, 1},\dots,W_{1, n_1}, W_{2, 1},\dots,W_{2, n_2}]^{T}$ to identify important words by multiplying the masks with input word embeddings,
\begin{equation}
\label{eq:mask_output}
\vec{\tilde{x}}=\vec{W}\odot \vec{x},
\end{equation}
where $\odot$ is an element-wise multiplication, the masked input $\vec{\tilde{x}} = [\vec{\tilde{x}}_1, \vec{\tilde{x}}_2]^{T}$, $\vec{\tilde{x}}_{i, j} = W_{i, j} \cdot \vec{x}_{i, j}$ ($i \in \{1, 2\},~j \in \{1, \dots, n_{i}\}$), and $W_{i, j} \in \{0,1\}$ is a binary random variable with 1 and 0 indicating to select or mask out the word $\vec{x}_{i, j}$ respectively. To generate an effective explanation, the word masks $\vec{W}$ should have the following properties: (1) correctly selecting important words for the model prediction; (2) removing as many irrelevant words as possible to keep the explanation concise; (3) selecting or masking out correlated words together from the sentence pair.

Previous work on learning individual word masks only focuses on the first two properties \citep{chen2020learning, de2020decisions}. To satisfy the third property, We propose \ourmethod to implicitly detect word correlations and distribute the correlated words into a group (e.g. \texttt{electric}, \texttt{guitar}, and \texttt{banjo} are assigned to $G1$ or $G2$ in \autoref{fig:illustrations}), and learn a group mask for these words. Specifically, we decompose each $W_{i,j}$ in $\vec{W}$ into two random variables,
\begin{equation}
  \label{eq:decom}
  W_{i,j}=\sum_{\iota=1}^{t}\delta(Z_{i,j},\iota)\delta(G,\iota),
\end{equation}
where $t$ is the predefined number of groups, and we will introduce how to pick up a $t$ in \autoref{subsec:implement}. 
$Z_{i, j} \in \{1, \dots, t\}$ indicates the word $\vec{x}_{i, j}$ belonging to which group, and $G \in \{1, \dots, t\}$ indicates which group takes the mask value 1, which means all words in this group are selected as important words, while other words in the rest groups are masked out. $\delta(a,b)$ is the Delta function with $\delta(a,b)=1$ when $a=b$, and 0 otherwise.
The conditional dependency of $\vec{W}$, $\vec{Z}$, and $G$ can be represented as a graphical model \footnote{$\vec{Z} \rightarrow \vec{W} \leftarrow G$: $\vec{Z}$ and $G$ are dependent given $\vec{W}$.}. The problem of learning $\vec{W}$ is equivalent to learning $\vec{Z}$ and $G$, that is learning word distributions among the groups and group importance. According to $\delta(Z_{i,j},\iota)$ and $\delta(G,\iota)$, the word masks $\vec{W}$ will keep or mask out all words in group $\iota$, which satisfies the third property.

\subsection{Learning \ourmethod}
\label{subsec:gmask}
We formulate the problem of learning \ourmethod by optimizing the following objective in terms of the three properties,
\begin{equation}
\label{eq:gmask_obj_1}
\max_{\Phi, \Psi}\expect{p(y\mid \vec{x}, \vec{z}, \vec{g})}{} - \gamma_{1} \mathcal{L}_{\vec{Z}}  - \gamma_{2} \mathcal{L}_{G},
\end{equation}
where $\Phi$ and $\Psi$ are parameters of $\vec{Z}$ and $G$ respectively, and $\vec{z}$ and $\vec{g}$ are samples of $\vec{Z}$ and $G$ respectively. We denote $\mathcal{L}_{\vec{Z}}$ and $\mathcal{L}_{G}$ as regularizations on $\vec{Z}$ and $G$ respectively, which are applied to make the learned masks satisfy the required properties. We will introduce the two regularization terms subsequently. $\gamma_{1}, \gamma_{2} \in\mathbb{R}_{+}$ are coefficients.

Optimizing the first term in \autoref{eq:gmask_obj_1} is to make the word masks $\vec{W}$ satisfy the first property, that is the model outputs the same prediction on the selected words as on the whole text. Given $\vec{Z}$ and $G$, we have word masks $\vec{W}$, and multiply them with input word embeddings, and obtain the masked input $\vec{\tilde{x}}$ as in \autoref{eq:mask_output}. The model output on $\vec{\tilde{x}}$ is $\tilde{y} = f(\vec{\tilde{x}})$. If the masks correctly select important words, the predicted label on the selected words should be equal to that on the whole input text. We can optimize the first term by minimizing the cross entropy loss ($\mathcal{L}_{ce}(\cdot, \cdot)$) between $\tilde{y}$ and $y$. The objective \autoref{eq:gmask_obj_1} can be rewritten as
\begin{equation}
\label{eq:gmask_obj_2}
\min_{\Phi, \Psi}\mathcal{L}_{ce}(y, \tilde{y}) + \gamma_{1} \mathcal{L}_{\vec{Z}} + \gamma_{2} \mathcal{L}_{G}.
\end{equation}

The last two terms in the optimization objective are to make word masks satisfy the second and third properties. We regularize $\vec{Z}$ to encourage each group contains some words from different sentences. We regularize $G$ to ensure only one or few groups are identified as important (with relatively large probabilities). Optimizing the cross entropy loss with the two regularization terms can make the word masks select the important group of words, where the words are selected from the input sentence pair and are correlated.

\paragraph{Regularizations on $\vec{Z}$ and $G$}
As each $Z_{i, j}$ ($i \in \{1, 2\},~j \in \{1, \dots, n_{i}\}$) indicates a word belonging to a specific group, it follows categorical distribution with probabilities $[\phi_{i, j}(1), \dots, \phi_{i, j}(t)]$, where $t$ is the predefined number of groups, and $\phi_{i, j}(\iota)$ ($\iota \in \{1, \dots, t\}$) represents the probability of the word in group $\iota$. Then we denote the parameters of $\vec{Z}$ as $\Phi$,
\begin{align}
\Phi &= \begin{bmatrix}
\phi_{1, 1}(1) & \cdots & \phi_{1, 1}(t) \\
\vdots & \cdots & \vdots \\
\phi_{1, n_1}(1) & \cdots & \phi_{1, n_1}(t) \\
\phi_{2, 1}(1) & \cdots & \phi_{2, 1}(t) \\
\vdots & \cdots & \vdots \\
\phi_{2, n_2}(1) & \cdots & \phi_{2, n_2}(t) \\
\end{bmatrix}
\end{align}

To ensure that each group contains some words from both input sentences, and also avoid assigning a bunch of words into one group, we distribute the words in each sentence evenly among all groups. Then each group implicitly captures the words from different sentences. We can regularize $\vec{Z}$ to achieve this goal. We sum each column of $\Phi$ along the upper half rows and lower half rows respectively, and obtain two vectors by taking averages, $\vec{\phi}^{U} = \frac{1}{n_1}[\sum_{j=1}^{n_1} \phi_{1, j}(1), \dots, \sum_{j=1}^{n_1} \phi_{1, j}(t)]$, $\vec{\phi}^{L} = \frac{1}{n_2}[\sum_{j=1}^{n_2} \phi_{2, j}(1), \dots, \sum_{j=1}^{n_2} \phi_{2, j}(t)]$. Then $\vec{\phi}^{U}$ and $\vec{\phi}^{L}$ are the distributions of two discrete variables $Z^{U}$ and $Z^{L}$, which also represent the word distributions of the two input sentences among groups. To make the distributions of words even, we maximize the entropy of $Z^{U}$ and $Z^{L}$, and have
\begin{equation}
\label{eq:z_loss}
\mathcal{L}_{\vec{Z}} = -(H(Z^{U}) + H(Z^{L})),
\end{equation}
where $H(\cdot)$ is entropy.

$G \in \{1, \dots, t\}$ also follows categorical distribution with probabilities $\Psi = [\psi(1), \dots, \psi(t)]$, where $\psi(\iota)$ ($\iota \in \{1, \dots, t\}$) represents the probability of group $\iota$ being selected. According to the relation of $\vec{W}$, $\vec{Z}$, $G$ in \autoref{eq:decom}, the word masks only keep the words assigned to the selected group. To ensure one or few groups have relatively large probabilities to be selected, we regularize $G$ by minimizing its entropy, that is $\mathcal{L}_{G} = H(G)$. The final optimization objective is
\begin{equation}
\label{eq:gmask_obj_3}
\min_{\Phi, \Psi}\mathcal{L}_{ce}(y, \tilde{y}) - \gamma_{1} (H(Z^{U}) + H(Z^{L})) + \gamma_{2} H(G).
\end{equation}

\paragraph{Optimization via sampling}
We adopt a sampling based method to solve \autoref{eq:gmask_obj_3} by learning the parameters of $\vec{Z}$ and $G$ (i.e. $\{\Phi, \Psi\}$). As the right part of \autoref{fig:framework} shows, we sample a $\vec{z}$ from the categorical distributions of $\vec{Z}$, where each row $\vec{z}_{i, j}$ is a one-hot vector, indicating the word $\vec{x}_{i, j}$ assigned to a specific group. And we sample a $\vec{g}$ from the categorical distribution of $G$, which is a vertical one-hot vector, indicating the selected group. Then we obtain a sample of word masks by multiplying $\vec{z}$ and $\vec{g}$, i.e. $\vec{w} = \vec{z}\cdot\vec{g}$, where the mask values corresponding to the words in the selected group are 1, while the rest are 0. We apply the masks on the input word embeddings and optimize \autoref{eq:gmask_obj_3} via stochastic gradient descent.

There are two challenges of the learning process: discreteness and large variance. We apply the Gumbel-softmax trick \citep{jang2016categorical, maddison2016concrete} to address the discreteness of sampling from categorical distributions in backpropagation. See \autoref{sec:sampling} for the continuous differentiable approximation of Gumbel-softmax. We do the sampling multiple times in \autoref{subsec:implement} and generate a batch of masked inputs of the original input data to decrease the variance in probing the model, and train for multiple epochs until the learnable parameters $\{\Phi, \Psi\}$ reach stable values.

\paragraph{Weighted word attributions} 
After training, we learn the parameters of $\vec{Z}$, i.e. $\Phi$, where each element $\phi_{i, j}(\iota) \in (0, 1)$ ($i \in \{1, 2\},~j \in \{1, \dots, n_{i}\},~\iota \in \{1, \dots, t\}$) represents the probability of word $\vec{x}_{i, j}$ belonging to group $\iota$. We also learn the parameters of $G$, i.e. $\Psi$, where each element $\psi(\iota) \in (0, 1)$ represents the importance of group $\iota$. According to the definition of word masks $\vec{W}$ in \autoref{subsec:wmask}, we know that each mask variable $W_{i, j}$ follows Bernoulli distribution, and the probability of $W_{i, j}$ taking 1 is denoted as $\theta_{i, j}$. We can compute $\theta_{i, j}$ based on the relation of $W_{i, j}$, $Z_{i, j}$ and $G$ in \autoref{eq:decom}, that is 
\begin{equation}
\label{eq:word_attr}
\theta_{i, j} = \sum_{\iota=1}^{t}\phi_{i, j}(\iota)\psi(\iota).
\end{equation}
We can see that $\theta_{i, j}$ is the expectation of $W_{i, j}$, representing the weighted attribution of the word $\vec{x}_{i, j}$ to the model predicition. Hence, we have a set of weighted word attributions $\Theta = [\theta_{1, 1},\dots,\theta_{1, n_1}, \theta_{2, 1},\dots,\theta_{2, n_2}]^{T}$ for extracting important words as an explanation.

\paragraph{Complexity} For a set of $n$ words, computing interactions between all word pairs costs $O(n^2)$ and aggregating words step by step to form a tree structure even costs more time \citep{singh2018hierarchical, chen2020generating}. \ourmethod circumvents the feature interaction detection by learning word groups. The complexity is $O(nt + t)$, where $t$ is the number of groups and usually $t\ll n$ in practice.

\subsection{Implementation Specification}
\label{subsec:implement}

We initialize the parameters of all categorical distributions ($\{\Phi, \Psi\}$) with $\frac{1}{t}$, which means all words have the same importance and do not have any preference to be in a specific group at the start of training. To stabilize the learning process, we sample 100 - 1000 examples (depending on the model and datasets) and train at most 100 epochs until converge. The coefficients $\gamma_{1}$ and $\gamma_{2}$ are hyperparameters. We empirically found $\gamma_{1}=10$ and $\gamma_{2}=1$ work well in our experiments.

In our pilot experiments, we found that preliminarily filtering out some noisy or irrelevant words can help decrease the learnable parameters, hence accelerating the training process.
Specifically, we adopt a simple word mask method from \citep{chen2020learning} to select a set of individual words for an input sentence pair before running \ourmethod.
This simple method, denoted as \textsc{IMASK}, will learn individual word attributions as masks $\vec{R}=\{R_{i, j}\}_{i \in \{1, 2\},~j \in \{1, \dots, n_{i}\}} \in \{0, 1\}^{n_1+n_2}$ regardless any correlation.
Then, based on the expected values of $\vec{R}$, we preliminarily select top $k$ words for \ourmethod to further learn weighted word attributions.
Within these top $k$ words, assume  $k_1$ words from the first input text and $k_2$ words from the second text, then we will set the number of groups as $t=\min (k_1,k_2)$, so that at least one group contains words from both sentences.
$k$ is a hyper-parameter associated with the average length of input texts. In the experiments, we set $k=10$.
Note that, the \textsc{IMASK} method adopted here can also be used as a baseline method for comparison.

\section{Experimental Setup}
\label{sec:setup}

We evaluate \ourmethod with two kinds of neural network models, decomposable attention model (DAttn) \citep{parikh2016decomposable} and BERT~\cite{devlin2018bert}, for two types of sentence pair modeling tasks on four datasets. We compare our method with four baselines.

\paragraph{Datasets} 
 e-SNLI \citep{camburu2018snli} is natural language inference task, where the model predicts the semantic relationship between two input sentences as entailment, contradiction, or neutral. Quora \citep{wang2017bilateral}, QQP \citep{wang2018glue} and MRPC \citep{dolan2005automatically} are paraphrase identification tasks, where the model decides whether two input texts are semantically equivalent or not. The statistics of the four datasets are in \autoref{sec:sup_exp}.

\paragraph{Models}
We adopt the decomposable attention model (DAttn) \citep{parikh2016decomposable} and BERT~\cite{devlin2018bert} model, and fine-tune the models on each downstream task to achieve the best performance, as \autoref{tab:acc} shows. The test results on QQP and MRPC are scored by the GLUE benchmark \footnote{\url{https://gluebenchmark.com/}{}}. The corresponding validation accuracy for each reported test accuracy is in \autoref{sec:val_acc}

\begin{table}
	\centering
	\begin{tabular}{ccccc}
		\toprule
		Models & e-SNLI & Quora & QQP & MRPC \\
		\midrule
		DAttn & 86.62 & 86.78 & 85.00 &  68.30 \\
		\addlinespace[0.2cm]
		BERT & 90.38 & 90.48 & 89.00 & 83.70  \\
		\bottomrule
	\end{tabular}
	\caption{The prediction accuracy (\%) of different models on the four datasets.}
	\label{tab:acc}
\end{table}

\paragraph{Baselines}
We compare \ourmethod with four baseline methods: (1) LIME \citep{ribeiro2016should}-fitting a local linear model with perturbations to approximate the neural network and produce word attributions; (2) L2X \citep{chen2018learning}-constructing a network to learn feature attributions by maximizing the mutual information between the selected features and model output; (3) IBA (\textit{Per-Sample} framework) \citep{schulz2020restricting} - learning feature attributions by optimizing the information bottleneck which restricts feature information flow by adding noise; (4) IMASK (\autoref{subsec:implement})-learning individual word masks. Note that here we use standalone IMASK as one of the baselines, as oppose to applying it for selecting preliminary important words for \ourmethod as in \autoref{subsec:implement}. 

More details about experimental setup are in \autoref{sec:sup_exp}, including data pre-processing and model configurations.

\section{Results and Discussion}
\label{sec:results}
We compare the faithfulness of generated post-hoc explanations via both quantitative and qualitative evaluations.


\begin{table}[t] 
 	\small
	\centering
	\begin{tabular}{p{0.7cm}p{1cm}p{1cm}p{0.7cm}p{0.7cm}p{0.9cm}}
		\toprule
		Models & Methods & e-SNLI & Quora & QQP & MRPC \\
		\midrule
		\multirow{4}{*}{DAttn} & LIME & 0.286 & 0.120 & 0.079 & 0.064  \\
		& L2X & 0.299 & 0.128 & 0.079 & 0.035  \\
		& IBA  & 0.354 & 0.137 & \textbf{0.104} & \textbf{0.109}  \\
		& \textsc{IMASK}  & 0.324 & 0.140 & 0.087 & 0.064  \\
		& \ourmethod  & \textbf{0.361} & \textbf{0.142} & 0.095 & 0.091  \\
		\midrule
		\multirow{4}{*}{BERT} & LIME & 0.221 & 0.153 & 0.110 & 0.062  \\
		& L2X  & 0.310 & 0.119 & 0.134 & 0.083  \\
		& IBA & 0.282 & 0.199 & 0.144 & 0.114  \\
		& \textsc{IMASK} & 0.292 & 0.232 & 0.139 & 0.130  \\
		&\ourmethod  & \textbf{0.319} & \textbf{0.309} & \textbf{0.181} & \textbf{0.200} \\
		\bottomrule
	\end{tabular}
	\caption{AOPC scores of different explanation methods with the DAttn and BERT models on the four datasets.}
	\label{tab:aopc_ave}
\end{table}

\subsection{Quantitative Evaluation}
\label{subsec:quanti_eva}
We adopt three metrics from prior work to evaluate the faithfulness of learned feature attributions: AOPC score \citep{nguyen2018comparing,samek2016evaluating}, post-hoc accuracy \citep{chen2018learning, chen2020learning}, and degradation score \citep{ancn17, schulz2020restricting}. 
We evaluate explanations on all test data for the MRPC dataset, and on 2000 examples randomly selected from the test set for other three datasets due to computational complexity. The average runtime is in \autoref{sec:run_time}.

\subsubsection{AOPC score}
\label{subsubsec:aopc}
We adopt the area over the perturbation curve (AOPC) \citep{nguyen2018comparing,samek2016evaluating} metric to evaluate the comprehensiveness of explanations to models. It calculates the average change of prediction probability on the predicted class over all examples by removing top $1\ldots u$ words in explanations.
\begin{equation}
\label{eq:aopc}
\text{AOPC}=\frac{1}{U+1}\langle\sum_{u=1}^Up(y|\vec{x})-p(y|\vec{x}_{\backslash1\ldots u})\rangle_{\vec{x}},
\end{equation}
where $p(y|\vec{x}_{\backslash1\ldots u})$ is the probability for the predicted class when words $1\ldots u$ are removed and $\langle\cdot \rangle_{\vec{x}}$ denotes the average over all test examples.
Higher AOPC score indicates better explanations.

\begin{figure*}[htb]
	\centering
	\subfigure[DAttn on e-SNLI]{
		\label{fig:dec_esnli}
		\includegraphics[width=0.21\textwidth]{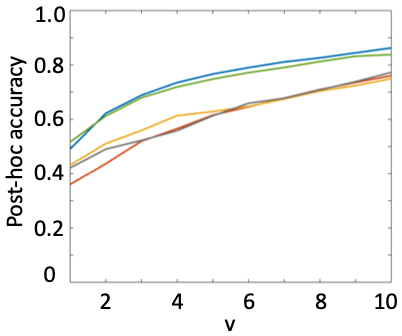}}
	\subfigure[DAttn on Quora]{
		\label{fig:dec_quora}
		\includegraphics[width=0.2\textwidth]{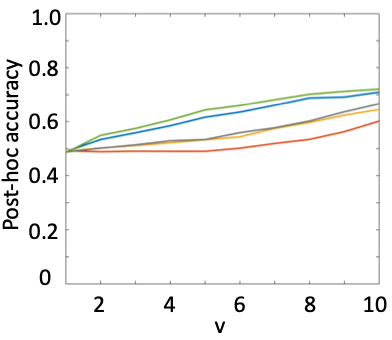}}
	\subfigure[DAttn on QQP]{
		\label{fig:dec_qqp}
		\includegraphics[width=0.2\textwidth]{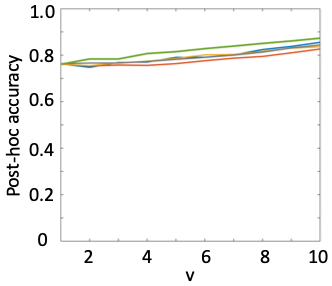}}
	\subfigure[DAttn on MRPC]{
		\label{fig:dec_mrpc}
		\includegraphics[width=0.21\textwidth]{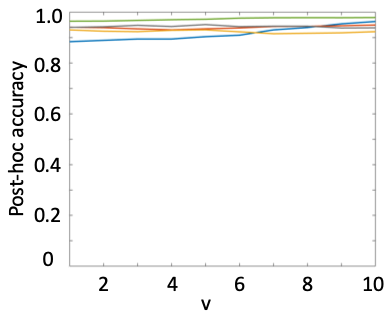}}
	\subfigure[BERT on e-SNLI]{
		\label{fig:bert_esnli}
		\includegraphics[width=0.2\textwidth]{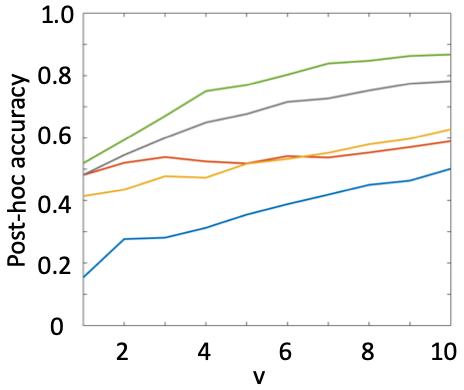}}
	\subfigure[BERT on Quora]{
		\label{fig:bert_quora}
		\includegraphics[width=0.21\textwidth]{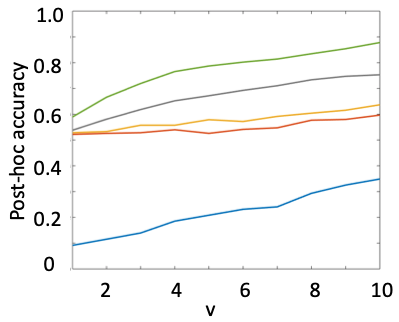}}
	\subfigure[BERT on QQP]{
		\label{fig:bert_qqp}
		\includegraphics[width=0.2\textwidth]{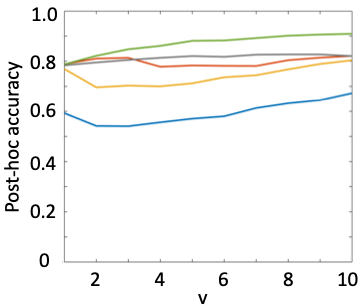}}
	\subfigure[BERT on MRPC]{
		\label{fig:bert_mrpc}
		\includegraphics[width=0.28\textwidth]{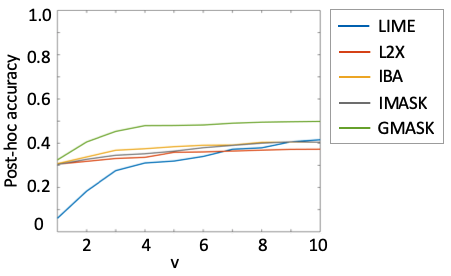}}
	\caption{Post-hoc accuracy of different explanation methods with the DAttn and BERT models on the four datasets.}
	\label{fig:post_hoc_acc}
\end{figure*}
\begin{figure*}[htb]
	\centering
	\subfigure[LIME]{
		\label{fig:lime_deg}
		\includegraphics[width=0.18\textwidth]{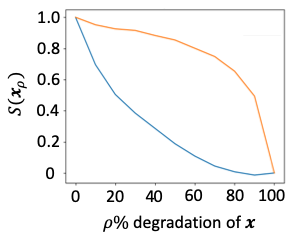}}
	\subfigure[L2X]{
		\label{fig:l2x_deg}
		\includegraphics[width=0.16\textwidth]{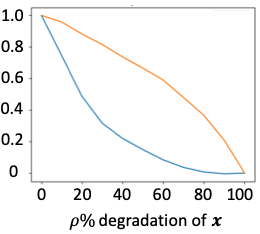}}
	\subfigure[IBA]{
		\label{fig:iba_deg}
		\includegraphics[width=0.16\textwidth]{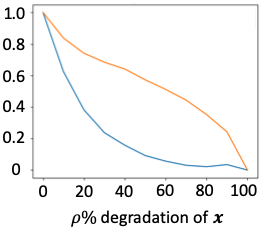}}
	\subfigure[\textsc{IMASK}]{
		\label{fig:IMASK_deg}
		\includegraphics[width=0.16\textwidth]{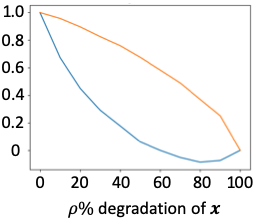}}
	\subfigure[\ourmethod]{
		\label{fig:gmask_deg}
		\includegraphics[width=0.22\textwidth]{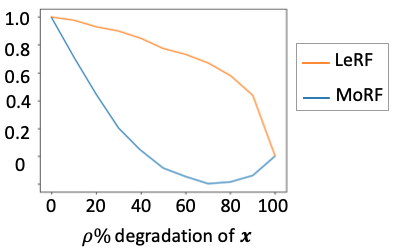}}
	\caption{Degradation test of different explanation methods with the DAttn model on the e-SNLI dataset.}
	\label{fig:degradation}
\end{figure*}

\autoref{tab:aopc_ave} shows the results of AOPC scores of different explanation methods when $U=10$. \ourmethod outperforms other baseline methods on most of the datasets. Especially for the BERT model, \ourmethod achieves significantly higher AOPC scores than other methods, indicating that BERT tends to rely on word correlations to make predictions. IBA and \textsc{IMASK}, either learning continuous or binary individual word masks, perform better than learning word attributions via an additional network (L2X) or using linear approximation (LIME).

\subsubsection{Post-hoc Accuracy}
\label{subsubsec:post_hoc_acc}

The post-hoc accuracy \citep{chen2018learning, chen2020learning} evaluates the sufficiency of important words to the model prediction. For each test data, we select top $v$ important words based on word attributions for the model to make a prediction, and compare it with the original prediction made on the whole input text. We compute the post-hoc accuracy on $M$ examples,
\begin{equation*}
\label{eq:post-hoc-acc}
\text{post-hoc-acc}(v)=\frac{1}{M}\sum_{m=1}^M\mathbbm{1} [y_{v}^{(m)}=y^{(m)}],
\end{equation*}
where $y^{(m)}$ is the predicted label on the m-th test data, and $y_{v}^{(m)}$ is the predicted label based on the top $v$ important words.
Higher post-hoc accuracy indicates better explanations.

\autoref{fig:post_hoc_acc} shows the results of post-hoc accuracy of different explanation methods where we increase $v$ from 1 to 10. Similar to the results of the AOPC scores, \ourmethod achieves higher post-hoc accuracy than other methods for both DAttn and BERT models. 

The explanations of \ourmethod for the BERT model achieve about $80\%$ post-hoc accuracy on all datasets except the MRPC dataset. This is only by relying on top 4 important words, which means that \ourmethod captures informative words for model predictions. The post-hoc accuracies of the BERT model on the MRPC dataset are lower than those on other three datasets because the average sentence length of MRPC is twice as long as the others, indicating that BERT tends to use larger context for predictions. The post-hoc accuracies of the DAttn model on the MRPC dataset are extremely high for all the explanation methods. The reason is that the prediction accuracy of DAttn model on the MRPC dataset is relatively low (\autoref{tab:acc}). 
Any random words picked up by explanations could make the model output wrong predictions since the original predictions on the whole texts are also wrong, hence causing high post-hoc accuracy.

\begin{table*}
  \small
  \centering
  \begin{tabular}{ccp{0.78\textwidth}}
    \toprule
    Models & Methods & Texts \\
    \midrule
     \multirow{5}{*}{DAttn} & LIME & a man playing \hlc[pink!25]{an} \hlc[pink!50]{electric} \hlc[pink!75]{guitar} on \hlc[pink!100]{stage} .  a man playing banjo on the floor .  \\
     & L2X & a man playing an \hlc[pink!75]{electric} guitar on stage .  a man \hlc[cyan!25]{playing} \hlc[cyan!50]{banjo} on the \hlc[cyan!100]{floor} . \\
    & IBA & a man playing an \hlc[pink!25]{electric} \hlc[pink!100]{guitar} on \hlc[pink!75]{stage} .  a man playing banjo on \hlc[cyan!25]{the} floor .  \\
     & \textsc{IMASK} & a man playing an electric \hlc[pink!50]{guitar} on stage .  \hlc[cyan!25]{a} \hlc[cyan!100]{man} playing \hlc[cyan!75]{banjo} on the floor .  \\
     & \ourmethod & a man playing an \hlc[pink!100]{electric} \hlc[pink!75]{guitar} on stage .  a man playing \hlc[cyan!25]{banjo} on the \hlc[cyan!50]{floor} . \\
    \rule{0pt}{4ex} 
    \multirow{5}{*}{BERT} & LIME & why are vikings \hlc[pink!50]{portrayed} wearing horned \hlc[pink!100]{helmets} ? why did vikings \hlc[cyan!25]{have} horns on their \hlc[cyan!75]{helmets} ?   \\
    & L2X & why \hlc[pink!100]{are} \hlc[pink!75]{vikings} portrayed wearing horned helmets ? \hlc[cyan!25]{why} \hlc[cyan!50]{did} vikings have horns on their helmets ?  \\
    & IBA & why \hlc[pink!100]{are} vikings portrayed wearing horned helmets ? \hlc[cyan!75]{why} did vikings have horns on their \hlc[cyan!25]{helmets} \hlc[cyan!50]{?}   \\
    & \textsc{IMASK} & why are vikings \hlc[pink!25]{portrayed} wearing horned helmets \hlc[pink!75]{?} why did vikings have \hlc[cyan!50]{horns} on their \hlc[cyan!100]{helmets} ?  \\
    & \ourmethod & why are vikings portrayed wearing \hlc[pink!75]{horned} helmets ? why did vikings \hlc[cyan!25]{have} \hlc[cyan!100]{horns} on their \hlc[cyan!50]{helmets} ?   \\
    \bottomrule
  \end{tabular}
  \caption{Examples of different explanations, where the top four important words are highlighted. The important words in the first and second sentences are highlighted in pink and blue colors respectively. The color saturation indicates word attribution. The first example is from the e-SNLI dataset, and the DAttn model makes a correct prediction as \textsc{contradiction}. The second example is from the Quora dataset, and the BERT model makes a correct prediction as \textsc{paraphrases}.}
  \label{tab:examples}
\end{table*}

\begin{table}[tbh]
 	\small
	\centering
	\begin{tabular}{p{0.7cm}p{1cm}p{1cm}p{0.7cm}p{0.7cm}p{0.9cm}}
		\toprule
		Models & Methods & e-SNLI & Quora & QQP & MRPC \\
		\midrule
		\multirow{5}{*}{DAttn} & LIME & 0.502 & 0.070 & 0.091 & 1.367  \\
		& L2X & 0.366 & 0.002 & 0.036 & 1.779  \\
		& IBA  & 0.423 & 0.110 & 0.197 &  2.775 \\
		& \textsc{IMASK}  & 0.436 & 0.152 & 0.214 & 2.037  \\
		& \ourmethod  & \textbf{0.620} & \textbf{0.178} & \textbf{0.238} & \textbf{2.790}  \\
		\midrule
		\multirow{5}{*}{BERT} & LIME & 0.188 & 0.192 & 0.087 & 0.018  \\
		& L2X  & 0.303 & 0.168 & 0.173 & -0.003  \\
		& IBA  & 0.166 & 0.038 & 0.176 & 0.050  \\
		& \textsc{IMASK}  & 0.369 & 0.303 & 0.172 & 0.251  \\
		&\ourmethod  & \textbf{0.576} & \textbf{0.726} & \textbf{0.707} & \textbf{0.533}  \\
		\bottomrule
	\end{tabular}
	\caption{Degradation scores of different explanation methods with the DAttn and BERT models on the four datasets.}
	\label{tab:degradation}
\end{table}

\subsubsection{Degradation Test}
\label{subsubsec:degra_test}

Degradation test \citep{ancn17, schulz2020restricting} evaluates the ranking of importance by removing the most important words or least important words first, and observing model prediction probability drop on the predicted class.
We draw two curves as shown in \autoref{fig:degradation}, one with the most relevant words removed first (MoRF) and another one with the least relevant words removed first (LeRF). x-axis is the percentage of words removed (degradation proportion), and y-axis is the normalized model output probability as
\begin{equation}
\label{eq:norm_y}
S(\vec{x}_{\rho}) = \frac{p(y|\vec{x}_{\rho}) - p(y|\vec{x}_{o})}{p(y|\vec{x}) - p(y|\vec{x}_{o})},
\end{equation}
where $\vec{x}$ is the original input, $y$ is the predicted label, $\vec{x}_{\rho}$ means $\rho\% (\rho \in [0, 100])$ degradation of $\vec{x}$, and $\vec{x}_{o}$ is full degradation. We compute the averages of $p(y|\vec{x}_{\rho})$, $p(y|\vec{x})$, and $p(y|\vec{x}_{o})$ over all test examples. The degradation score is calculated as the integral between the MoRF and LeRF curves,
\begin{equation}
\label{eq:deg_score}
\text{degra-score} = \int_{\rho=0}^{100} \frac{S^{L}(\vec{x}_{\rho}) - S^{M}(\vec{x}_{\rho})}{100} d\rho,
\end{equation}
where $S^{L}(\vec{x}_{\rho})$ and $S^{M}(\vec{x}_{\rho})$ are normalized model outputs by removing the least or most important words respectively.
Higher degradation score is better.

\autoref{tab:degradation} shows the results of degradation scores of different explanation methods. 
\ourmethod shows superiority to other baseline methods under this metric. \autoref{fig:degradation} shows the degradation test results of DAttn model on the e-SNLI dataset.
\ourmethod can distinguish both important and unimportant words, while IBA does not learn the correct order of unimportant words. LIME does not perform well in identifying important words, but captures the correct order of unimportant words. The MoRF and LeRF curves of L2X and \textsc{IMASK} are relatively symmetric, but not as expanded as \ourmethod.

\subsection{Qualitative Evaluation}
\label{subsec:quali_eva}
\autoref{tab:examples} shows different explanations on two examples from e-SNLI and Quora respectively. See \autoref{sec:extra_exp} for more examples. For the first example, the DAttn model makes a correct prediction as \textsc{contradiction}. For the second example, the BERT model also makes a correct prediction as \textsc{paraphrases}. We highlight the top four important words, where the words in the first and second sentences are in pink and blue colors respectively. The color saturation indicates word attribution.

For the first example, LIME and IBA mainly capture the important words from the first sentence, while ignoring the ones in the second sentence (e.g. \texttt{banjo}, \texttt{floor}). On the contrary, L2X focuses on the words in the second sentence, while ignoring the important word \texttt{guitar} in the first sentence. \textsc{IMASK} picks up two irrelevant words \texttt{man} and \texttt{a} as important words, which can not explain the model prediction. \ourmethod correctly identifies top four important words and captures two correlated words \texttt{guitar} and \texttt{banjo} from the two input sentences respectively.

For the second example, only \ourmethod captures the two important correlated words \texttt{horned} and \texttt{horns}, which explains why the BERT model predicts the two input questions as paraphrases. LIME captures the overlapped word \texttt{helmets} in the two sentences, while L2X only selects some irrelevant words. Both IBA and \textsc{IMASK} identify a question mark as the important word, which is untrustworthy to the model prediction.

\section{Conclusion}
\label{sec:conclusion}
In this paper, we focused on sentence pair modeling and proposed an effective method, \ourmethod, learning group masks for correlated words and calculating weighted word attributions. We tested \ourmethod with two different neural network models on four datasets, and assessed its effectiveness via both quantitative and qualitative evaluations.

\section{Ethical Considerations}
\label{sec:ethical}

The motivation of this work is aligned with the merits of explainable AI, in which the goal is to increase the trustworthiness of neural network models in decision making.
One potential ethical concern of this work is that explanations can be used to design adversarial examples for attacking.
Although the main focus of this work is about generating faithful explanations, we do realize the importance of whether human users can actually understand explanations.
To address this concern, a better strategy is to collaborate with HCI experts in our future work.
In addition, we provide necessary implementation details to make sure the results in this paper are reproducible.

\bibliographystyle{acl_natbib}
\bibliography{ref}

\clearpage
\newpage
\appendix
\section{Sampling with Gumbel-softmax Trick}
\label{sec:sampling}
For a random variable $A \in \{1, \dots, t\}$ which follows categorical distribution with probabilities $[\lambda_1, \dots, \lambda_t]$. We draw samples from a Gumbel(0, 1) distribution for each category $\iota \in\{1, \dots, t\}$:
\begin{equation}
\label{eq:gumbel_softmax1}
s_{\iota}=-\log(-\log u),~u\sim \text{Uniform}(0, 1),
\end{equation}
and then apply a temperature-dependent softmax over the $t$ categories with each dimension calculated as
\begin{equation}
\label{eq:gumbel_softmax2}
a_{s_{\iota}}=\frac{\exp((\log(\lambda_{\iota})+s_{\iota})/\tau)}{\sum_{\iota}\exp((\log(\lambda_{\iota})+s_{\iota})/\tau)},
\end{equation}
where $\tau$ is a hyperparameter called the softmax temperature.

\section{Supplement of Experiment Setup}
\label{sec:sup_exp}
\begin{table*}
  \centering
  \begin{tabular}{ccccccc}
    	\toprule
         Datasets & \textit{C} & \textit{L} & \textit{V} & \textit{\#train} & \textit{\#dev} & \textit{\#test} \\
         \midrule
         e-SNLI & 3 & 10.2 & 64291 & 549K & 9K & 9K  \\
         Quora & 2 & 11.5 & 85249 & 384K & 10K & 10K  \\
         QQP & 2 & 11.1 & 126266 & 364K & 40K & 391K  \\
         MRPC & 2 & 22 & 15547 & 3668 & 408  & 1725  \\
         \bottomrule
  \end{tabular}
  \caption{Summary statistics for the datasets, where \textit{C} is the number of classes, \textit{L} is average sentence length, \textit{V} is vocab size, and \textit{\#} counts the number of examples in the \textit{train/dev/test} sets.}
  \label{tab:datasets}
\end{table*}

\paragraph{Datasets} The datasets are all in English. \autoref{tab:datasets} shows the statistics of the four datasets. We adopt the data splits of e-SNLI \citep{camburu2018snli} from the ERASER benchmark \footnote{\url{https://www.eraserbenchmark.com/}{}}. We adopt the data splits of Quora  released by \citet{wang2017bilateral}. The data splits of QQP \citep{wang2018glue} and MRPC \citep{dolan2005automatically} are from the GLUE benchmark. 
We clean up the text by converting all characters to lowercase, removing extra whitespaces and special characters, and build vocabulary.

\paragraph{Models}
We set the hidden size of feed forward networks in the DAttn model \citep{parikh2016decomposable} as 300, and initialize word embeddings with pre-trained \texttt{fastText} \citep{bojanowski2017enriching}. For BERT model~\cite{devlin2018bert}, we use the pretrained BERT-base model\footnote{\url{https://github.com/huggingface/pytorch-transformers}{}} with 12 transformer layers, 12 self-attention heads, and the hidden size of 768.

We implement the models in PyTorch 3.6. The number of parameters in the DAttn and BERT models are 11046303 and 109484547 respectively. We fine-tune hyperparameters manually for each model to achieve the best prediction accuracy, such as learning rate $lr \in \{1e-4, 1e-3,\cdots, 1\}$, clipping norm $clip \in \{1e-3, 1e-2, \cdots, 1, 5, 10\}$.

\section{Validation Performance}
\label{sec:val_acc}
The corresponding validation accuracy for each reported test accuracy is in \autoref{tab:val_acc}.

\begin{table*}
	\centering
	\begin{tabular}{lllll}
		\toprule
		Models & e-SNLI & Quora & QQP & MRPC \\
		\midrule
		DAttn & 87.75 & 87.36 & 87.19 & 73.77 \\
		\addlinespace[0.2cm]
		BERT & 90.43 & 91.21 & 91.31 & 86.52  \\
		\bottomrule
	\end{tabular}
	\caption{The validation accuracy (\%) of different models on the four datasets.}
	\label{tab:val_acc}
\end{table*}

\section{Average Runtime}
\label{sec:run_time}
The average runtime of each approach for each model on each dataset is recorded in \autoref{tab:run_time}. All experiments were performed on a single NVidia GTX 1080 GPU. Note that L2X is efficient in generating explanations for test data, but it costs more time on training the interpretation model on the whole training set.

\begin{table*}[tbh]
	\centering
	\begin{tabular}{cccccc}
		\toprule
		Models & Methods & e-SNLI & Quora & QQP & MRPC \\
		\midrule
		\multirow{4}{*}{DAttn} & LIME & 20.17 & 20.45 & 20.43 & 19.12  \\
		& L2X & 0.01 & 0.02 & 0.02 & 0.01  \\
		& IBA  & 9.87 & 9.93 & 9.96 & 9.24  \\
		& \textsc{IMASK}  & 0.17 & 0.19 & 0.18 & 0.12  \\
		& \ourmethod  & 11.10 & 11.23 & 11.25 &  10.96 \\
		\midrule
		\multirow{4}{*}{BERT} & LIME & 13.41 & 14.01 & 14.08 & 13.21  \\
		& L2X  & 0.02 & 0.02 & 0.02 & 0.01  \\
		& IBA & 3.29 & 3.38 & 3.34 & 3.20  \\
		& \textsc{IMASK}  & 0.62 & 0.70 & 0.71 & 0.59  \\
		&\ourmethod & 43.24 & 43.56 & 43.77 & 42.84  \\
		\bottomrule
	\end{tabular}
	\caption{The average runtime (s/example) of each approach for each model on each dataset.}
	\label{tab:run_time}
\end{table*}

\section{Examples of Different Explanations}
\label{sec:extra_exp}

\autoref{tab:sec:extra_exp} shows more examples of different explanations for the DAttn and BERT model on different datasets.

\begin{table*}
  \small
  \centering
  \begin{tabular}{ccp{0.6\textwidth}}
    \toprule
    Model/Dataset/Prediction & Methods & Texts \\
    \midrule
     \multirow{5}{*}{DAttn/Quora/\textsc{paraphrases}} & LIME & who are \hlc[pink!25]{some} \hlc[pink!100]{famous} \hlc[pink!75]{nihilists} ?  what would a \hlc[cyan!50]{nihilistic} president do to the us ?  \\
     & L2X & who are some famous nihilists ?  what would a nihilistic president \hlc[cyan!50]{do} \hlc[cyan!25]{to} \hlc[cyan!75]{the} \hlc[cyan!100]{us} ?  \\
    & IBA & who \hlc[pink!50]{are} some \hlc[pink!100]{famous} \hlc[pink!75]{nihilists} ?  what \hlc[cyan!25]{would} a nihilistic president do to the us ?  \\
     & \textsc{IMASK} & \hlc[pink!100]{who} are some famous nihilists ?  \hlc[cyan!50]{what} would a nihilistic president \hlc[cyan!75]{do} \hlc[cyan!25]{to} the us ?  \\
     & \ourmethod & who are some famous \hlc[pink!50]{nihilists} ?  what \hlc[cyan!25]{would} a \hlc[cyan!100]{nihilistic} \hlc[cyan!75]{president} do to the us ?  \\
     
    \rule{0pt}{4ex} 
    
    \multirow{5}{*}{DAttn/QQP/\textsc{nonparaphrases}} & LIME & can \hlc[pink!50]{i} register shares of a \hlc[pink!100]{private} limited company in india ?  can a \hlc[cyan!25]{school} in india be registered as a \hlc[cyan!75]{private} limited company ?  \\
     & L2X & can i register shares of a private limited \hlc[pink!100]{company} in india ?  can a school in \hlc[cyan!25]{india} be registered as a private \hlc[cyan!50]{limited} \hlc[cyan!75]{company} ? \\
    & IBA & can i register shares of a private limited \hlc[pink!100]{company} in \hlc[pink!75]{india} ?  can a \hlc[cyan!50]{school} in india be registered \hlc[cyan!25]{as} a private limited company ?  \\
     & \textsc{IMASK} & can i register shares of \hlc[pink!75]{a} private limited company \hlc[pink!100]{in} india ?  can a school \hlc[cyan!25]{in} india be registered \hlc[cyan!50]{as} a private limited company ?  \\
     & \ourmethod & can i \hlc[pink!75]{register} shares of a private limited \hlc[pink!100]{company} in india ?  can a \hlc[cyan!50]{school} in india be registered as a \hlc[cyan!25]{private} limited company ? \\
     
    \rule{0pt}{4ex} 
    
    \multirow{5}{*}{DAttn/MRPC/\textsc{paraphrases}} & LIME & these documents are indecipherable to me and the \hlc[pink!50]{fact} is that \hlc[pink!100]{this} investigation has led nowhere the lawyer said .  these documents are indecipherable to me the lawyers said and the \hlc[cyan!25]{fact} is that \hlc[cyan!75]{this} investigation has led nowhere .  \\
     & L2X & these documents are indecipherable to \hlc[pink!100]{me} and the fact \hlc[pink!50]{is} that this investigation has led nowhere the lawyer said .  these documents are indecipherable to \hlc[cyan!75]{me} the lawyers said and the fact \hlc[cyan!25]{is} that this investigation has led nowhere . \\
    & IBA & these documents are indecipherable \hlc[pink!50]{to} me and the fact \hlc[pink!25]{is} that this investigation has led nowhere the lawyer said .  these documents \hlc[cyan!75]{are} indecipherable to me the lawyers said and the fact is that this \hlc[cyan!100]{investigation} has led nowhere .  \\
     & \textsc{IMASK} & these documents \hlc[pink!75]{are} indecipherable to me and the fact \hlc[pink!100]{is} that this investigation has led nowhere the lawyer \hlc[pink!50]{said} .  these documents are indecipherable \hlc[cyan!25]{to} me the lawyers said and the fact is that this investigation has led nowhere .  \\
     & \ourmethod & these documents are indecipherable to me and the fact is that this \hlc[pink!75]{investigation} has led \hlc[pink!25]{nowhere} the lawyer said .  these documents are indecipherable \hlc[cyan!50]{to} me the lawyers said and the fact is that this \hlc[cyan!100]{investigation} has led nowhere . \\
     
    \rule{0pt}{4ex} 
    
    \multirow{5}{*}{BERT/e-SNLI/\textsc{contradiction}} & LIME & a band singing and playing electric guitar \hlc[pink!25]{for} a \hlc[pink!100]{crowd} of people \hlc[pink!75]{.}  the band is backstage \hlc[cyan!50]{.}  \\
     & L2X & a band \hlc[pink!50]{singing} and playing electric guitar for a crowd of people \hlc[pink!25]{.}  the band \hlc[cyan!75]{is} \hlc[cyan!100]{backstage} . \\
    & IBA & a band singing and playing electric guitar \hlc[pink!25]{for} \hlc[pink!100]{a} \hlc[pink!50]{crowd} of people .  \hlc[cyan!75]{the} band is backstage . \\
     & \textsc{IMASK} & \hlc[pink!50]{a} band \hlc[cyan!25]{singing} and playing \hlc[pink!100]{electric} guitar for a \hlc[pink!75]{crowd} of people .  the band is backstage . \\
     & \ourmethod & a band singing and playing \hlc[pink!25]{electric} guitar for a \hlc[pink!75]{crowd} of people .  \hlc[pink!50]{the} band is \hlc[cyan!100]{backstage} . \\
     
    \rule{0pt}{4ex} 
    
    \multirow{5}{*}{BERT/QQP/\textsc{paraphrases}} & LIME & how \hlc[pink!100]{do} i quit smoking ? how \hlc[cyan!75]{do} i give up \hlc[cyan!50]{on} \hlc[cyan!25]{cigarette} smoking ?  \\
     & L2X & how do i quit smoking \hlc[pink!100]{?} how do i give up \hlc[cyan!50]{on} cigarette \hlc[cyan!25]{smoking} \hlc[cyan!75]{?} \\
    & IBA & how do i \hlc[pink!25]{quit} \hlc[pink!100]{smoking} ? how do i give up on \hlc[cyan!75]{cigarette} \hlc[cyan!50]{smoking} ?  \\
     & \textsc{IMASK} & how \hlc[pink!50]{do} i quit smoking ? how do i \hlc[cyan!100]{give} \hlc[cyan!25]{up} \hlc[cyan!75]{on} cigarette smoking ?  \\
     & \ourmethod & how do \hlc[pink!75]{i} \hlc[pink!100]{quit} smoking ? how do i \hlc[cyan!50]{give} \hlc[cyan!25]{up} on cigarette smoking ? \\
     
    \rule{0pt}{4ex} 
    
    \multirow{5}{*}{BERT/MRPC/\textsc{nonparaphrases}} & LIME & mgm , nbc and liberty executives were not \hlc[pink!25]{immediately} \hlc[pink!75]{available} for comment . \hlc[cyan!100]{a} microsoft spokesman was not \hlc[cyan!50]{immediately} available to comment .   \\
    & L2X & mgm , nbc and liberty executives were not immediately available for comment . a microsoft \hlc[cyan!75]{spokesman} was not immediately available \hlc[cyan!100]{to} \hlc[cyan!25]{comment} \hlc[cyan!50]{.}  \\
    & IBA & \hlc[pink!75]{mgm} , \hlc[pink!100]{nbc} and \hlc[pink!50]{liberty} executives were not immediately available for comment . a microsoft \hlc[cyan!25]{spokesman} was not immediately available to comment .   \\
    & \textsc{IMASK} & mgm , nbc and liberty executives were \hlc[pink!25]{not} \hlc[pink!50]{immediately} available \hlc[pink!100]{for} comment . a microsoft \hlc[cyan!75]{spokesman} was not immediately available to comment . \\
    & \ourmethod & mgm , nbc and \hlc[pink!25]{liberty} executives were not \hlc[pink!50]{immediately} available for comment . a microsoft spokesman was \hlc[cyan!75]{not} \hlc[cyan!75]{immediately} available to comment .   \\
    \bottomrule
  \end{tabular}
  \caption{Examples of different explanations for the DAttn and BERT model on different datasets.}
  \label{tab:sec:extra_exp}
\end{table*}

\end{document}